\documentclass{article}
\usepackage{graphicx} 
\usepackage{fullpage}
\usepackage{hyperref}
\usepackage{booktabs}
\usepackage{xcolor}
\usepackage{multirow}
\usepackage{tcolorbox}
\usepackage{placeins}
\tcbuselibrary{listingsutf8}
\usepackage{amsmath}
\usepackage{amssymb}
\usepackage{authblk}

\usepackage{etoolbox}
\usepackage{hyperref}

\newcommand{\CXR}{\mathrm{CXR}}

\definecolor{skcolor}{rgb}{0.6,0.5,0.1}

\title{What Causes Postoperative Aspiration?}
\author[1]{Supriya Nagesh\thanks{Corresponding author: nsupriy@amazon.com}}
\author[2]{Karina Covarrubias}
\author[2]{Robert El-Kareh}
\author[1]{Shiva Prasad Kasiviswanathan}
\author[1]{Nina Mishra}
\affil[1]{Amazon}
\affil[2]{UC San Diego Health}

\date{}
\begin{document}

\maketitle

\begin{abstract}
{\bf \noindent Background:}
Aspiration, the inhalation of foreign material into the lungs, significantly impacts surgical patient morbidity and mortality. This study develops a machine learning (ML) model to predict postoperative aspiration, enabling timely preventative interventions.\\

{\bf \noindent Methods:}
From the MIMIC-IV database of over 400,000 hospital admissions, we identified 826 surgical patients (mean age: 62, 55.7\% male) who experienced aspiration within seven days post-surgery, along with a matched non-aspiration cohort. Three ML models: XGBoost, Multilayer Perceptron, and Random Forest were trained using pre-surgical hospitalization data to predict postoperative aspiration. To investigate causation, we estimated Average Treatment Effects (ATE) using Augmented Inverse Probability Weighting. \\

{\bf \noindent Results:}
Our ML model achieved an AUROC of 0.86 and 77.3\% sensitivity on a held-out test set. Maximum daily opioid dose, length of stay, and patient age emerged as the most important predictors. ATE analysis identified significant causative factors: opioids (0.25 ± 0.06) and operative site (neck: 0.20 ± 0.13, head: 0.19 ± 0.13). Despite equal surgery rates across genders, men were 1.5 times more likely to aspirate and received 27\% higher maximum daily opioid dosages compared to women.\\

{\bf \noindent Conclusions:}
ML models can effectively predict postoperative aspiration risk, enabling targeted preventative measures. Maximum daily opioid dosage and operative site significantly influence aspiration risk. The gender disparity in both opioid administration and aspiration rates warrants further investigation. These findings have important implications for improving postoperative care protocols and aspiration prevention strategies.

\end{abstract}

\section{Introduction}

Aspiration refers to the entry of foreign material from the oropharynx or gastrointestinal tract into the larynx and lower respiratory tract~\cite{mandell2019aspiration}. This can involve the inhalation of oropharyngeal secretions, gastric contents, or blood past the vocal cords and into the tracheobronchial tree. Symptoms of aspiration may include coughing, wheezing, dyspnea, tachypnea, and in severe cases, respiratory distress or respiratory failure~\cite{depaso1991aspiration,warner1993clinical}.  Aspiration can lead to serious health complications, including chemical pneumonitis, bacterial pneumonia, lung abscess formation, and acute respiratory distress syndrome~\cite{marik2001aspiration}. 
These complications can result in prolonged hospitalization, increased morbidity, and in severe cases, mortality~\cite{sparn2021risk}. 


Aspiration risk factors encompass a broad spectrum of conditions that compromise protective airway mechanisms and increase susceptibility to pulmonary complications. Primary predisposing factors include altered consciousness (due to medications, substance use, anesthesia, or seizures), neurological disorders causing dysphagia, and mechanical disruption of normal protective barriers~\cite{mandell2019aspiration}. Upper gastrointestinal tract disorders, particularly those affecting esophageal function or involving surgical intervention, significantly elevate aspiration risk~\cite{lee2018aspiration}. Iatrogenic factors such as tracheostomy, endotracheal intubation, and nasogastric feeding can compromise natural protective mechanisms~\cite{dibardino2015aspiration}. Poor oral hygiene represents a significant modifiable risk factor, potentially increasing both the bacterial load and pathogenicity of aspirated material~\cite{pace2010association,terpenning2001aspiration}.

Aspiration events are often unwitnessed~\cite{son2017pneumonitis}, particularly in clinical settings where patients may be unattended or unable to communicate effectively. This lack of direct observation poses a challenge in diagnosing aspiration promptly. To confirm an aspiration event, chest radiography is used to detect the presence of infiltrates in the lungs~\cite{mandell2019aspiration,depaso1991aspiration}. These infiltrates, which represent areas of increased density in the lung tissue, are a hallmark sign of aspiration and usually become visible on chest X-rays within two hours of the aspiration event~\cite{warner1993clinical}. 


The primary objective of this research is to develop a robust machine learning solution capable of predicting aspiration events following surgical procedures. Our focus is specifically on patients without a known history of aspiration, as this population presents a unique and challenging predictive scenario. This targeted approach is grounded in the understanding that patients with prior aspiration events have an elevated risk of recurrence~\cite{mandell2019aspiration,masuda2022risk}, making their cases more straightforward to predict. In contrast, identifying first-time aspiration events in previously unaffected individuals poses a more complex challenge. By concentrating on this subset of patients, we aim to address a critical gap in perioperative care and risk assessment. 

For patients identified as high-risk for aspiration, several preventative measures can be implemented to mitigate this risk. One of the primary interventions is maintaining patients in a semi-recumbent position, typically with the head of the bed elevated 30 to 45 degrees ~\cite{orozco1995semirecumbent,loeb2003interventions}, which can significantly reduce the likelihood of gastroesophageal reflux and subsequent aspiration. Enteral feeding strategies, such as post-pyloric feeding, can bypass the stomach and minimize the risk of regurgitation~\cite{marik2001aspiration,fox1995aspiration}. Speech and swallowing evaluations conducted by specialized therapists can identify specific deficits and guide individualized management plans~\cite{huang2023videofluoroscopy}. The use of thickened liquids and adoption of a chin-down position during swallowing can improve airway protection in patients with dysphagia~\cite{vilardell2016comparative,newman2016effect}. Additionally, proactive management of nausea and dysphagia through pharmacological and non-pharmacological means can further reduce aspiration risk~\cite{warusevitane2015safety,allami2022evaluation}. By implementing these evidence-based preventative measures in patients identified as high-risk through predictive modeling, healthcare providers can potentially reduce the incidence of postoperative aspiration and its associated complications.

A key contribution of this paper is the development of a machine learning algorithm that predicts postoperative aspiration. 
Prior efforts have developed predictive models for composite postoperative pulmonary complications~\cite{sabate2014predicting,miskovic2017postoperative,zhou2024predictive}. However, these composite outcomes contain a heterogeneous set of conditions that do not lend themselves to one clear risk mitigation strategy. We sought to create a model specifically predicting postoperative aspiration to enable early initiation of aspiration-preventing efforts. 
Similarly, previous studies have focused on predicting the risk of postoperative pulmonary complications~\cite{canet2010prediction}, predicting postoperative nausea and vomiting, which could lead to complications and prolonged hospital stay~\cite{zhou2023predicting}, whereas our work specifically targets postoperative aspiration. Unlike~\cite{park2022simple} and~\cite{bowles2017preoperative}, which develop risk models for aspiration following a specific procedure, our approach aims to predict aspiration risk across many types of surgeries.


Additionally, this study identifies key predictive features, confirming the importance of patient age~\cite{muder1998pneumonia,makhnevich2019aspiration,dibardino2015aspiration} and pre-operative hospital stay while highlighting opioid administration’s role in aspiration risk. Notably, the maximum Milligram Morphine Equivalent (MME) administered in a day is the strongest opioid-related predictor, emphasizing the association of pre-operative peak opioid dosage and aspiration risk. These insights offer actionable guidance for clinical practice. By highlighting the potential impact of high-dose opioid administration, our research underscores the importance of maximizing multimodal non-narcotic pain management strategies in the postoperative period, previously shown to decrease postoperative nausea, a known risk factor for aspiration~\cite{chiu2018improved,simpson2019pain}. 

Finally, our study examines the role of gender in postoperative aspiration, building upon and extending previous findings in this area~\cite{van2011risk,nativ2022predictors}. Consistent with prior research, our results confirm that men are also more likely to aspirate following surgery compared to women. However, our analysis reveals a novel and important nuance to this gender disparity. We find that among patients who do aspirate, men are administered higher maximum daily doses of opioids. Specifically, the median of the maximum daily opioid given to men is 25 times that given to women who aspirate. This striking difference in opioid administration between genders could raise important questions about pain management practices and their potential contribution to aspiration risk. It suggests that the higher incidence of aspiration in men may not solely be due to physiological differences but could also be influenced by disparities in pain management approaches. 

\section{Methods}\label{sec:methods}

\paragraph{Dataset.}

\begin{figure}[h]
    \centering
    \includegraphics[width=1\textwidth]{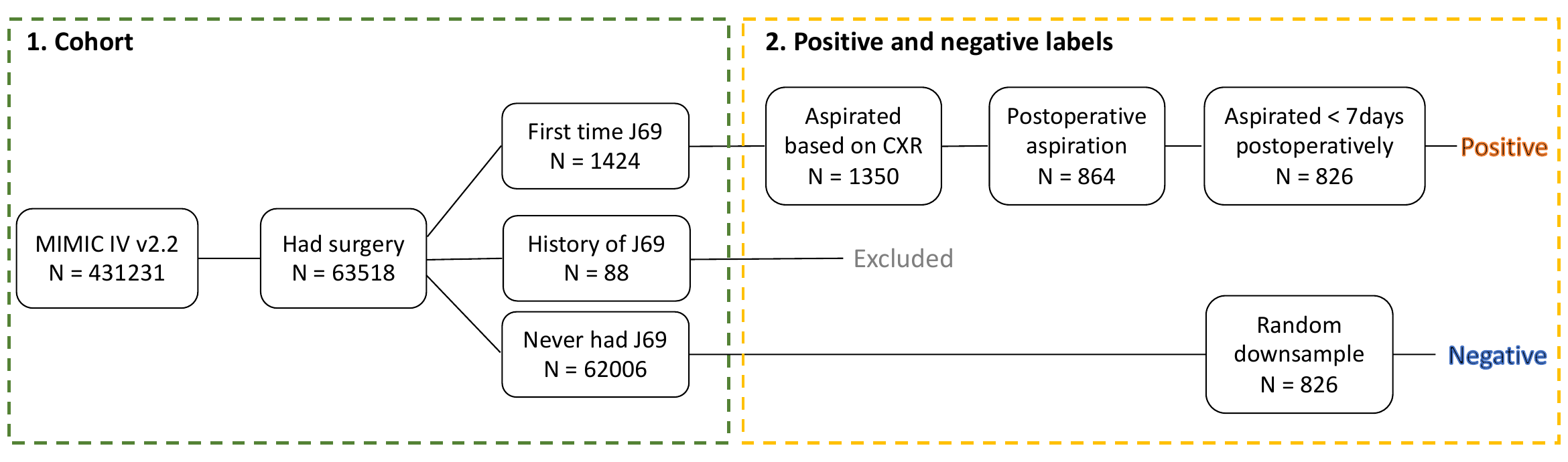}
    \caption{{\bf Flowchart of cohort selection.} Our goal is to build a predictive model for postoperative aspiration using a cohort of positive and negative samples from MIMIC-IV. Of about 400K admissions, roughly 15\% involved surgery. Admissions were labeled positive if the patient had no prior J69 ICD code, a chest X-ray confirming aspiration, and the X-ray occurred within 7 days of surgery. Patients with a history of aspiration were excluded to avoid an easier prediction task. To balance classes, negatives were downsampled to match the number of positives.}
    \label{fig:flowchart}
\end{figure}

The MIMIC-IV v2.2 dataset, a de-identified repository from Beth Israel Deaconess Medical Center (2008-2019), \cite{mimiciv,goldberger2000physiobank}, forms the basis for our findings.  The hospital-wide EHR database is used, with cohort selection focused on postoperative aspiration. Figure~\ref{fig:flowchart} outlines the selection process. Patients undergoing surgery during admission are included, while those with prior aspiration (J69 ICD code in previous MIMIC-IV admissions) are excluded to ensure a more challenging and clinically relevant prediction task. Though this may lead to false exclusions due to coding errors, it prevents the more critical issue of including such patients.

Designing a predictive model for postoperative aspiration requires data on patients who aspirated (with timing) and those who did not. A key challenge is obtaining reliable labels, as aspiration is often unwitnessed. While ICD codes exist, they may be inaccurate~\cite{campbell2001systematic} since they are assigned by chart reviewers who are not directly involved in patient care. 
To address this, aspiration events are confirmed through chest X-ray reports, using imaging time as a proxy for aspiration time. Chest X-ray reports are processed with a large language model (Claude 3.0 Sonnet~\cite{anthropic2024claude3}) to confirm evidence of aspiration.  Physionet responsible use guidelines are followed in the use of LLMs~\cite{responsibleUse}.
The prompt for determining aspiration is in Appendix~\ref{sec:prompt}. 


\paragraph{Label and Feature Extraction.}

In order to determine which patients aspirated, 
a time window is needed to define ``postoperative'' -- should it be 1 day, 2 days or more, following surgery.  Figure~\ref{fig:postopdays} shows that the number of aspirations peaks on the day of surgery and monotonically declines thereafter.  A 7-day time window is chosen to define postoperative, since $>95\%$ of aspirations in our dataset occur within 7 days of surgery. This study aims to predict postoperative aspiration within this period, though the approach is adaptable to other time frames.

\begin{figure}[h]
    \centering
    \includegraphics[width=0.9\linewidth, trim=0 1cm 0 1cm, clip]{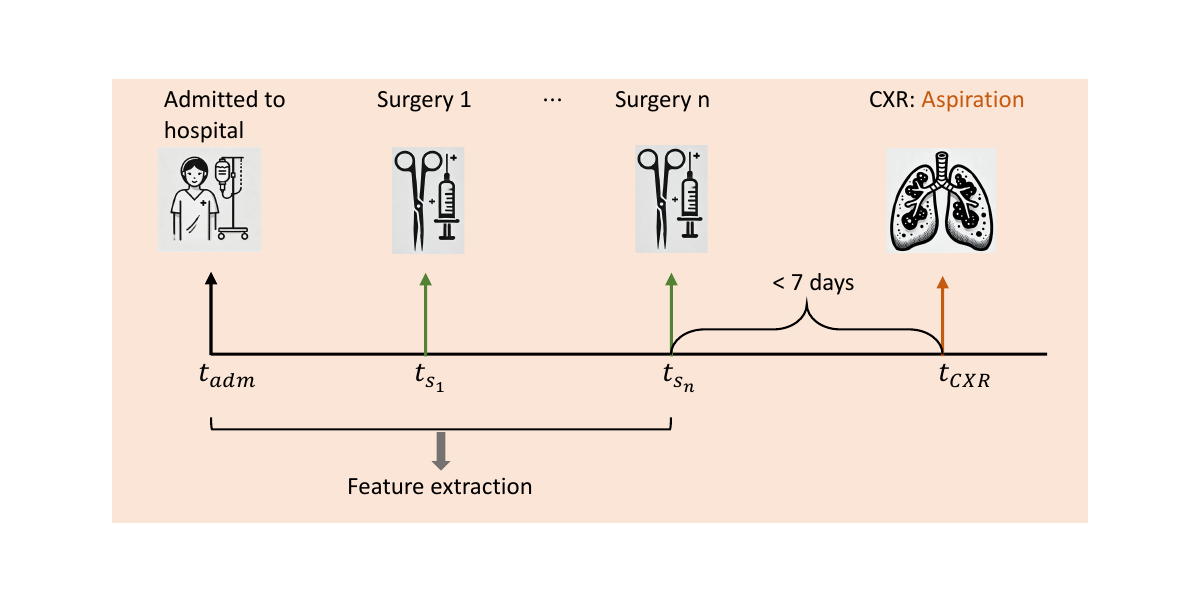}
    \caption{Feature selection for a positive case. Extracting data from admission to the last surgery before aspiration (or last surgery for non-aspirating patients) to predict aspiration within 7 days post-operation. }
    \label{fig:pt_timeline}
\end{figure}

In our cohort, patients often undergo multiple surgeries per admission. 
A stylized example is shown in Figure~\ref{fig:pt_timeline}. This patient was admitted to the hospital at time $t_{adm}$ and underwent $n$ surgeries at times $t_1, t_2, \cdot \cdot \cdot , t_{s_n}$. Aspiration for this patient was confirmed through a chest X-ray performed at time $t_{\CXR}$, which is used as a proxy for aspiration time. To ensure a clear predictive window, surgeries after $t_{\CXR}$ are excluded, making $t_{s_n}$ the last surgery before aspiration. 

The goal is to predict aspiration using pre-operative data from the time of admission ($t_{adm}$) up to the last surgery $t_{s_n}$.  Pre-operative data is used to predict post-operative aspiration.  The rationale is that a surgeon can use a prediction at the time of the last surgery $t_{s_n}$ to take suitable aspiration precautionary measures.

Features that were used to train the model include demographics, comorbidities, medications, and surgical details (full list in Table~\ref{tab:features}). For aspirating patients, features are extracted up to  time $t_{s_n}$; for non-aspirating patients, up to their last recorded surgery. This ensures that predictions rely only on preoperative data and maintain consistency across groups. 

\begin{table}[h!]
\centering
\begin{tabular}{c c }
\toprule
\textbf{Category} & \textbf{Features} \\
\toprule
\multirow{2}{*}{\textbf{Demographics}} & Age, Gender, \\
                                       & Language, Race \\
                                       \hline
\multirow{1}{*}{\textbf{Current admission}}    & Length of stay until surgery \\
\hline
\multirow{3}{*}{\textbf{Condition}}    & History of stroke, History of dyslipidemia, \\    
                                       & History of dysphagia, Obesity, \\
                                       & Hypertension, Diabetes \\

\hline
\multirow{3}{*}{\textbf{Type of surgery}} & Head, Neck, Spine, Thorax, \\
                                          & Upper abdomen, Lower abdomen, Pelvis,\\
                                          & Upper limbs, Lower limbs, Skin\\
\hline
\multirow{2}{*}{\textbf{Medications}}  & Opioids, Antiemetics, Analgesics, \\
                                       & Antidiabetic medications \\
\bottomrule
\end{tabular}
\caption{Features used to build a predictive model.}
\label{tab:features}

\end{table}

\paragraph{Predictive Modeling.}
The task is framed as a binary classification problem: predicting whether a patient will experience aspiration within seven days post-operation. Three models are evaluated: multi-layer perceptron (MLP),  XGBoost classifier, and Random Forest. These models were selected for their complementary advantages: MLP captures complex patterns using neural network architectures, while XGBoost and Random Forest, as ensemble methods, offer robustness and strong performance on tabular data.

The dataset was split 70\%-30\% for training and testing to ensure sufficient data for model learning and evaluation. Final performance was assessed on the held-out test set. Models were implemented using \texttt{scikit-learn}, with hyperparameter details in Appendix~\ref{sec:hyp}.

\paragraph{Causal Factors: Average Treatment Effect.}
\label{sec:ate}

This study examines the measurable causal impact of specific medications or operative sites on postoperative aspiration. The Average Treatment Effect (ATE), which measures the causal effect of a treatment or intervention on an outcome, is computed and averaged across the population.  In order to formally define ATE, there are two potential outcomes: let $Y_i(1)$ be the outcome variable representing whether the patient $i$ aspirated if they were treated, and $Y_i(0)$ be the outcome variable representing whether the patient $i$ aspirated if they were not treated. Formally, ATE is defined as 
$$\text{ATE} := \mathbb{E}[Y_i(1) - Y_i(0)] = E[Y(1) - Y(0)],$$ where the average is over the individuals $i$. It is well-known that ATE is identifiable under standard causal assumptions: unconfoundedness, positivity, consistency, and no interference~\cite[Theorem 2.1]{neal2020introduction}. Under these assumptions, ATE can be identified as:
\begin{equation*}
    \text{ATE} =\mathbb{E}_{X}[\mathbb{E}[Y | T=1, X] - \mathbb{E}[Y | T=0, X]],
\end{equation*}
where
\begin{itemize}
    \item $Y$: The outcome variable, representing whether the patient aspirated (binary: 1 for aspiration, 0 otherwise).
    \item $T$: The treatment variable, representing whether the patient was exposed to a specific medication (e.g., opioids, analgesics) or underwent a particular type of surgery (e.g., neck surgery).
    \item $X$: Covariates or patient features, including demographics, comorbidities, medications, and surgery details.
\end{itemize}

ATE is computed using the Augmented Inverse Probability Weighting (AIPW) method, which combines propensity score weighting and outcome regression to ensure robust estimation. First, a propensity score model predicts the probability of receiving treatment, and outcome regression models predict potential outcomes for treated and control groups based on covariates. A Decision Tree Classifier is used for estimating the propensity score, and Decision Tree Regressors are used for predicting potential outcomes. The AIPW estimator integrates outcome predictions and weighted residuals, ensuring doubly robust estimation: ATE is consistent if either the propensity score or outcome model is correctly specified. Confidence intervals for the estimate are calculated using bootstrap resampling. Details of the hyperparameter selections are provided in Appendix~\ref{sec:hyp}. 

ATE was computed for medications including Opioids, Non-opioid Analgesics (Acetaminophen), Insulin, Antiemetics (Ondansetron), and Saline flush. 
These medications were chosen because opioids likely increase aspiration risk due to their effects on respiratory depression~\cite{shook1990differential}, sedation~\cite{goodman1996goodman}, and impaired airway reflexes~\cite{tomazini2018effects}. In contrast, non-opioid analgesics, saline flushes, insulin, and antiemetics lack significant central nervous system depressant effects, making them less likely to contribute to aspiration.

\paragraph{Conditional Average Treatment Effect (CATE).}
The conditional average treatment effect (CATE) is the most standard parameter in heterogeneous effect estimation. Formally, CATE is defined as: $$\text{CATE}: = \mathbb{E}[Y(1) - Y(0) | X=x],$$  
measuring the expected difference in outcomes had those with covariates $X = x$ been treated versus not. As with ATE, under standard causal assumptions, CATE can be identified as, 
\begin{equation*}
    \text{CATE} =  \mathbb{E}[Y | T=1, X=x] - \mathbb{E}[Y | T=0, X=x]].
\end{equation*}



CATE is estimated using the AIPW framework on gender-specific subgroups, applying the same models as in ATE estimation. Confidence intervals are computed via bootstrap resampling to ensure robust estimates of treatment heterogeneity.

\section{Results}

\paragraph{Predicting Aspiration.}

The Random Forest classifier outperformed other models, achieving 77.4\% accuracy and an AUROC of 0.86 (Table~\ref{tab:model_accuracies} shows results of the three models). The confusion matrix (Figure~\ref{fig:confmat}) shows a sensitivity of 77.3\% and specificity of 77.5\%, highlighting its effectiveness in predicting postoperative aspiration. A threshold of 0.5 was used for classification, which is a standard choice. However, this threshold can be adjusted to prioritize higher sensitivity or specificity, depending on clinical requirements. 



Feature importance scores from the Random Forest classifier reveal key predictors of aspiration risk, measured by mean decrease in impurity (Gini importance). The top 15 features (Figure~\ref{fig:feat_imp}) include maximum daily opioid dosage, hospital stay length before surgery, age, and surgical location (thorax, upper abdomen). These findings align with~\cite{canet2010prediction}, which links thoracic and upper abdominal surgeries to pulmonary complications. Understanding these risk factors may aid clinical decision-making by identifying high-risk patients for closer monitoring.

\paragraph{Causal Factors.}

\begin{figure}[h]
    \centering
    \includegraphics[width=0.6\linewidth]{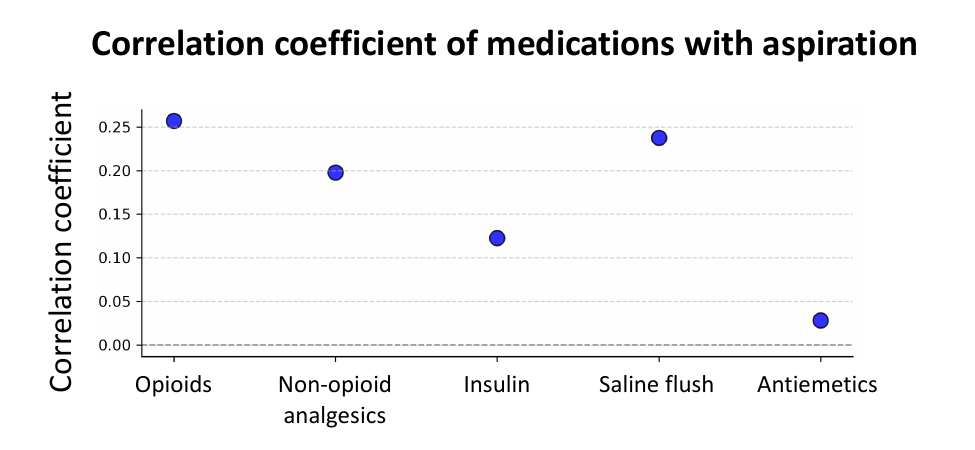}
    \caption{Correlation between different medications and postoperative aspiration. We see a higher correlation between opioids and saline flush with aspiration}
    \label{fig:asp_corr_meds}
\end{figure}

A correlation was observed between maximum daily opioid dosage (MME) and aspiration, as well as between saline flush dosage and aspiration, as illustrated in Figure~\ref{fig:asp_corr_meds}. To further investigate, we analyze the ATE of various medications (Table~\ref{tab:ate}). Opioids show the highest ATE ($0.25 \pm 0.06$), indicating a 25\%  increased aspiration risk (with a margin of error of ±6\%) after adjusting for covariates. Other medications, such as non-opioid analgesics and saline flushes, have smaller ATE values with wider confidence intervals, showing no strong causal effect. 

\begin{table}[h]
\centering
\begin{tabular}{lcc}
\toprule
\textbf{Medication} & \textbf{Average Treatment Effect} \\
\midrule
\textbf{Opioids} &  $\mathbf{0.25 \pm 0.06}$\\
Non-opioid analgesics &   $0.06 \pm 0.20$\\
Insulin & $0.01 \pm 0.21$\\
Saline flush &  $0.17 \pm 0.31$\\
Antiemetics & $0.02 \pm 0.25$\\
\bottomrule
\end{tabular}
\caption{ATE values on aspiration for different medications. Opioids have a statistically significant effect on aspiration with $p < 0.05$. None of the other medications have a significant effect on aspiration.}
\label{tab:ate}

\end{table}

\begin{figure}[h]
    \centering
    \includegraphics[width=0.8\textwidth]{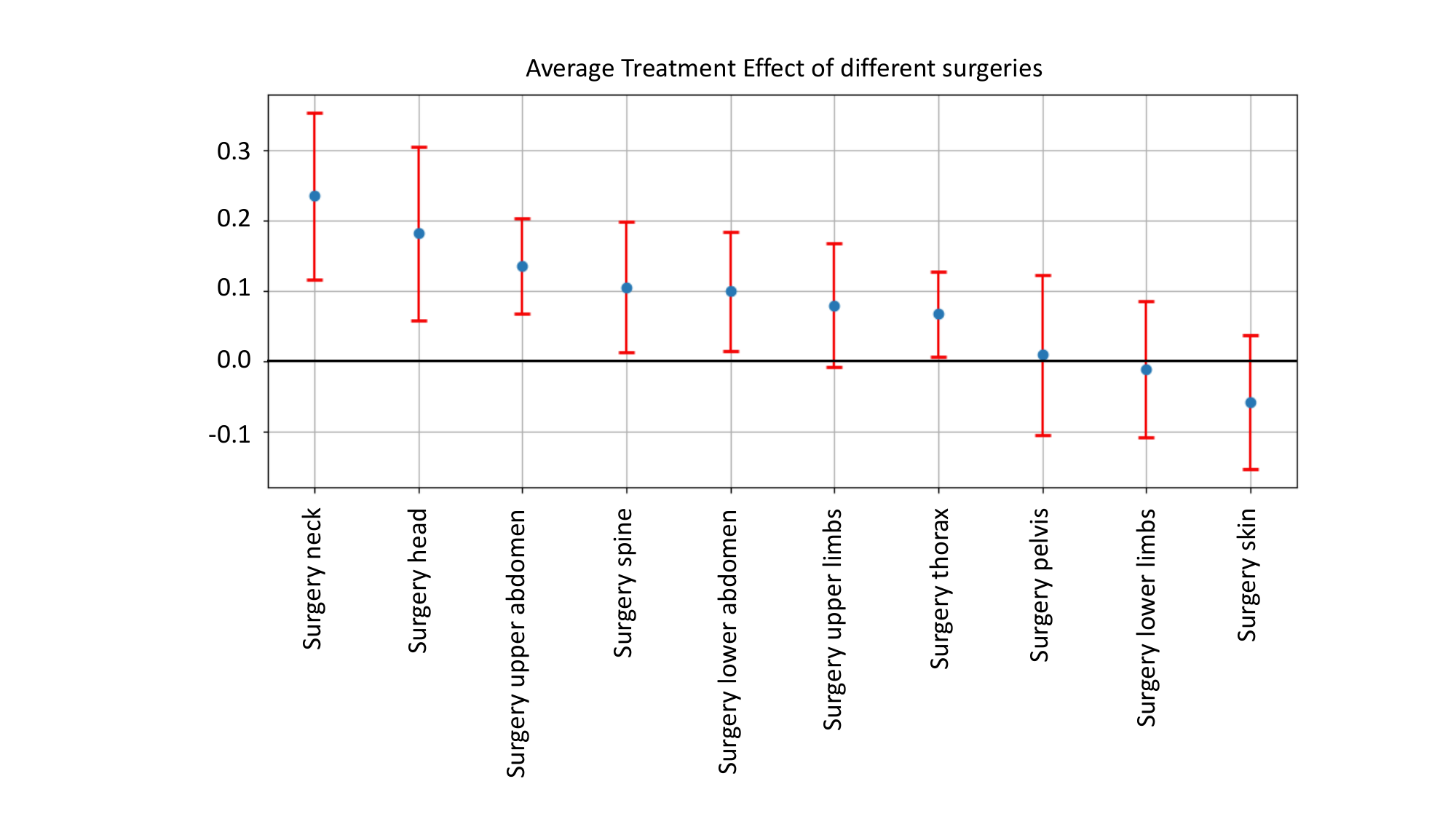}
    \caption{ATE of operative sites on postoperative aspiration, showing significant effects for neck, head, upper abdomen, and spine surgeries ($p < 0.05$)}
    \label{fig:surgery-type-ate}
\end{figure}

The impact of operative sites on aspiration risk was observed, with ATE values shown in Figure~\ref{fig:surgery-type-ate}. Surgeries involving the neck and head had the highest causal effect, indicating a significantly increased risk within seven days post-operation, while sites like skin or limbs showed minimal impact.
These findings align with the physiological impact of surgeries in high-risk regions, where structural and functional changes contribute to increased aspiration susceptibility. Surgeries involving the neck, head, and upper abdomen~\cite{canet2010prediction} heighten aspiration risk by affecting airway protection and swallowing. 


The high ATE value for opioids indicates that these medications substantially increase the likelihood of postoperative aspiration, emphasizing the importance of careful opioid management in patients at risk. Similarly, the elevated ATE values for neck and head surgeries highlight the need for targeted preoperative planning and postoperative care in these cases. 

\paragraph{Demographic Differences.}

Since prior work reports gender differences in aspiration rates~\cite{van2011risk,nativ2022predictors}, the next results investigate gender differences in postoperative aspiration in the MIMIC-IV dataset.  The key findings are that: men do also aspirate more often than women in this dataset, men do receive higher opioid doses, and there is no statistically significant difference in conditional average treatment effect of opioids among the genders.

\begin{figure}[h]
    \centering
    \includegraphics[width=0.7\linewidth]{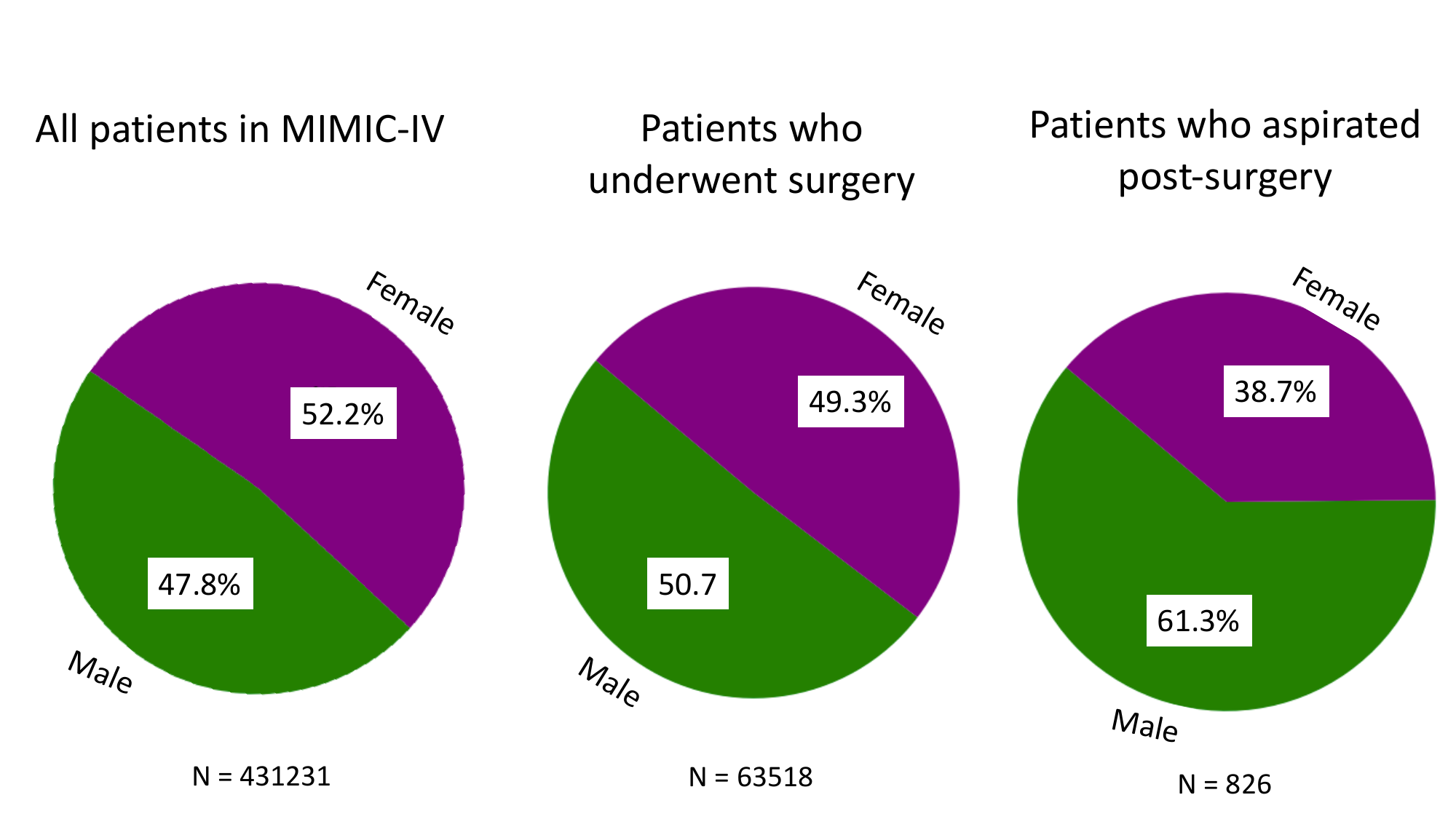}
    \caption{Difference in the distribution based on gender. Both MIMIC IV and patients who underwent surgeries were about 50\% male-female. However, among the patients who aspirated post-operation, 61.3\% of them were male.}
    \label{fig:asp_gender}
\end{figure}

In terms of gender distribution, Figure~\ref{fig:asp_gender} shows that while the MIMIC-IV database has a nearly equal gender distribution, 61.3\% of post-aspiration cases are male, nearly twice the female percentage (38.7\%), suggesting higher risk. Figure~\ref{fig:opioids_gender} further shows that males receive higher opioid doses, with a median max daily MME 25 times that of females. These findings align with our earlier reported results linking higher opioid dosages to increased aspiration risk.

In terms of conditional average treatment effect, the CATE of opioids is $0.23\pm0.07$  for males and $0.26\pm0.1$ for females. No statistically significant difference of treatment effect exists between the two genders, i.e., there is a consistent treatment effect of opioids across genders.

\paragraph{Limitations}

Several important limitations should be considered when interpreting our findings. First, our analysis is limited to a single Boston-based clinical dataset, raising concerns about generalizability. While large, its applicability to other populations is uncertain. Validation with diverse datasets is needed to assess broader relevance.  Second, our reliance on ICD code J69 for aspiration cases may introduce selection bias due to coding inaccuracies. Additionally, using an LLM to interpret chest X-ray reports, while efficient, adds potential interpretation bias. Third, our study faces a temporal limitation, as aspiration events are often unwitnessed. We use the first confirmatory chest X-ray timestamp as a proxy, which may not precisely reflect the actual timing, impacting temporal analyses.
Lastly, our predictive model, designed for real-time risk assessment, excludes postoperative interventions. While this enhances point-of-care utility, it does not account for factors like patient positioning or medications that may impact aspiration risk, potentially affecting accuracy in such scenarios.

\section{Conclusion}

What causes patients to aspirate post-operatively? To the best of our knowledge, no prior work has studied average treatment effect or attempted to quantify causal factors. In this study, based on 400K hospital admissions and 826 surgeries, the average treatment effect of pre-operative opioids on post-operative aspiration is quantified and found to be (0.25 +/- 0.06). In addition, operative site, such as neck and head have an ATE of 0.20 +/. Moreover, this study finds that men are 1.5 times more likely to aspirate than women, and are given higher doses 27\% higher maximum daily opioid dosage.  The findings suggest that opioid dosing regimens ought to be reconsidered, particularly for men, and especially for head and neck surgeries.

A predictive model for postoperative aspiration is developed using surgery-time data from a public dataset. Results show that data-driven methods effectively predict aspiration risk and highlight key factors. Notably, opioid administered strongly correlates with increased aspiration risk, as indicated by high ATE, suggesting a need to revisit pain management protocols. Elevated ATE values for head, neck, and upper abdomen surgeries emphasize the need for tailored postoperative planning. These findings inform clinical decision-making to reduce aspirat    ion risk. Future work could validate these insights on diverse datasets and explore the causal effect of other drugs, such as GLP-1s~\cite{yeo2024increased}.
\section*{Data Availability}
The MIMIC-IV dataset is publicly available at 
{\tt https://physionet.org/content/mimiciv/2.2/} and the corresponding Chest X-Ray reports are available at {\tt https://physionet.org/content/mimic-cxr/2.1.0/}

\section*{Funding}
The study received no outside funding. 

\section*{Dedication}
To Monique Fox
\pagebreak

\bibliographystyle{alpha}
\bibliography{bib}
\makeatletter
\apptocmd{\appendix}{\renewcommand{\theHfigure}{appendix.\arabic{figure}}}{}{}
\makeatother
\section*{Appendix}

\appendix
\renewcommand{\thesection}{A\arabic{section}}
\renewcommand{\thefigure}{A.\arabic{figure}}
\renewcommand{\thetable}{A.\arabic{table}}
\setcounter{section}{1}
\setcounter{figure}{0}
\setcounter{table}{0}

\paragraph{Roadmap.}

This appendix is structured as follows. It begins with an overview of the basic demographic distributions in MIMIC-IV and our cohort. This is followed by a depiction of the machine learning methodology. Details on label and feature selection, along with the corresponding results, are then presented. To provide insight into the model’s errors, an analysis of false negatives is included. Finally, a section outlining the hyperparameters and software packages used in the experiments concludes the appendix.

\subsection{Patient Distribution}

\begin{figure}[h]
    \centering
    \includegraphics[width=0.8\linewidth,trim=50 50 50 50, clip]{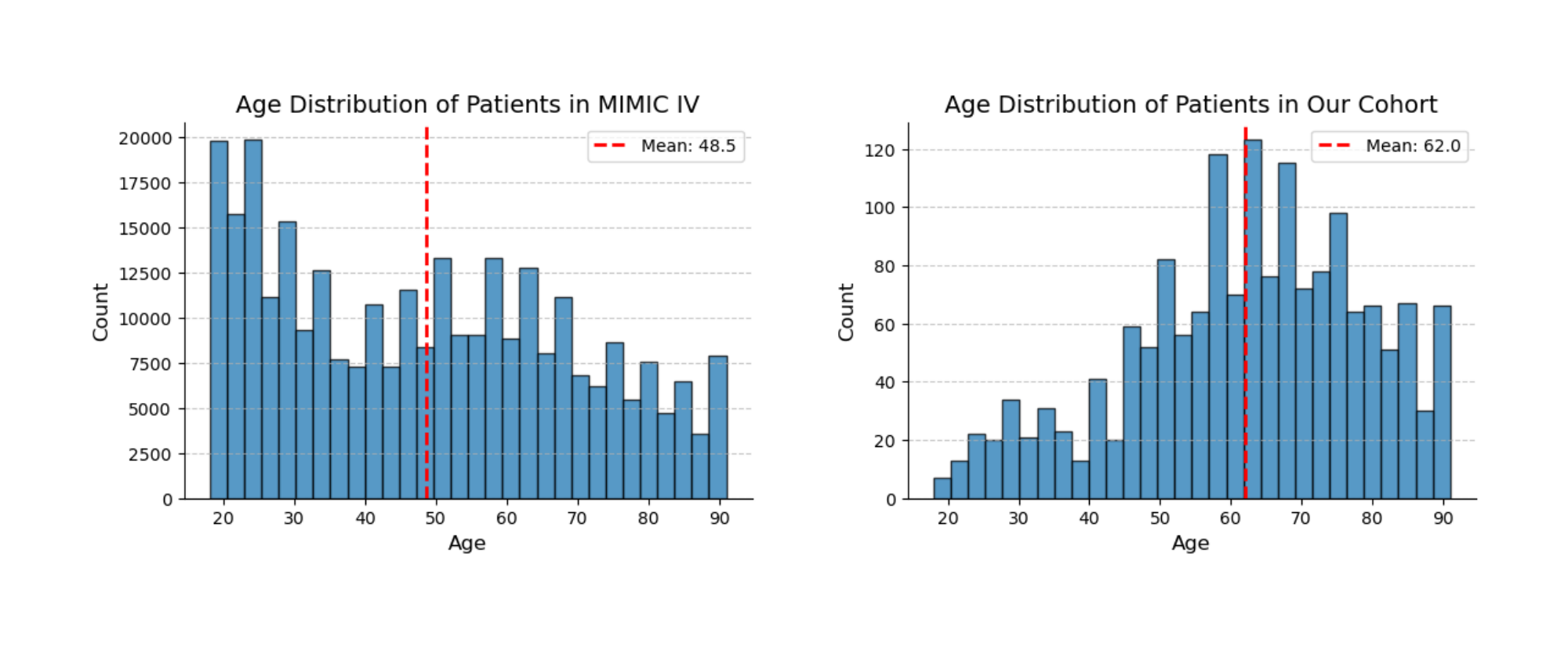}
    \caption{Distribution of patient age in MIMIC IV and our cohort.}
    \label{app:age_mimic_cohort}
\end{figure}

\begin{figure}[h]
    \centering
    \includegraphics[width=0.8\linewidth,trim=50 70 50 50, clip]{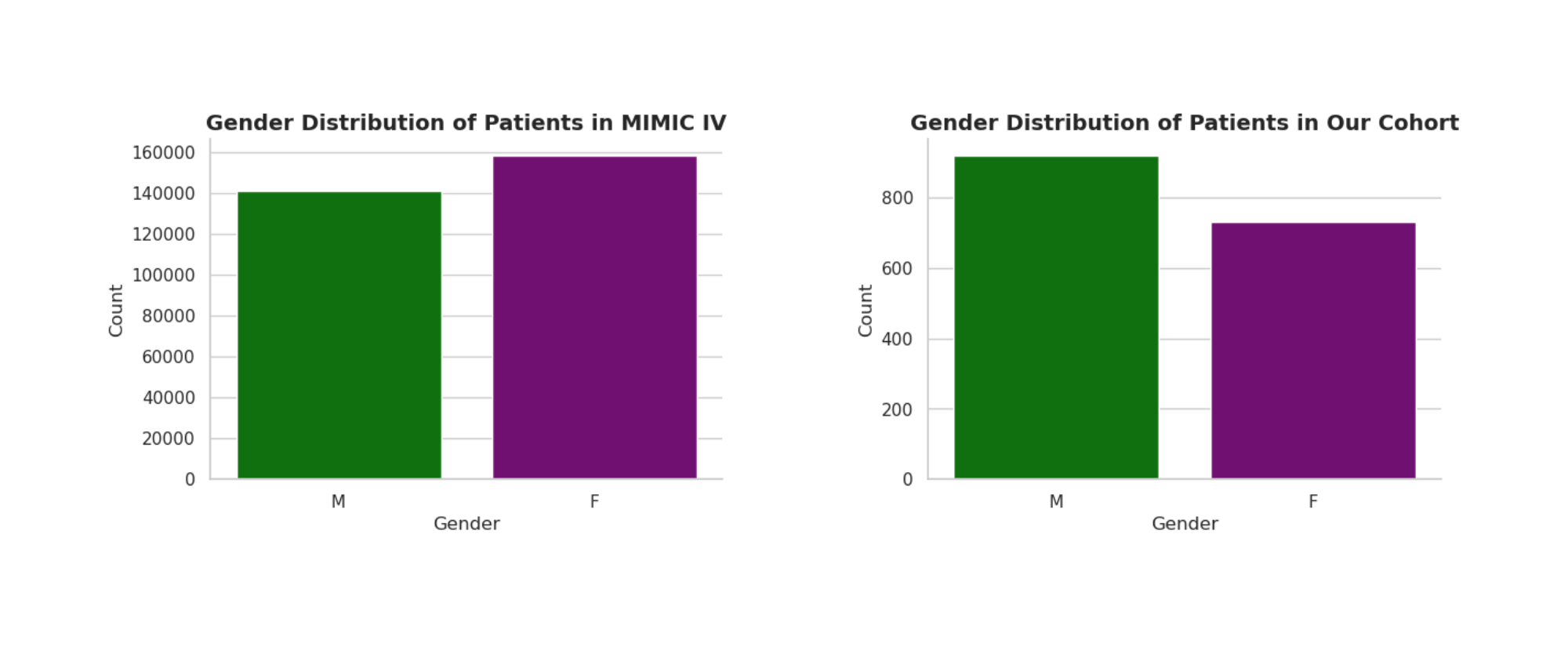}
    \caption{Distribution of patient gender in MIMIC IV and our cohort.}
    \label{app:gender_mimic_cohort}
\end{figure}

\begin{figure}[h]
    \centering
    \includegraphics[width=0.8\linewidth,trim=55 30 55 25, clip]{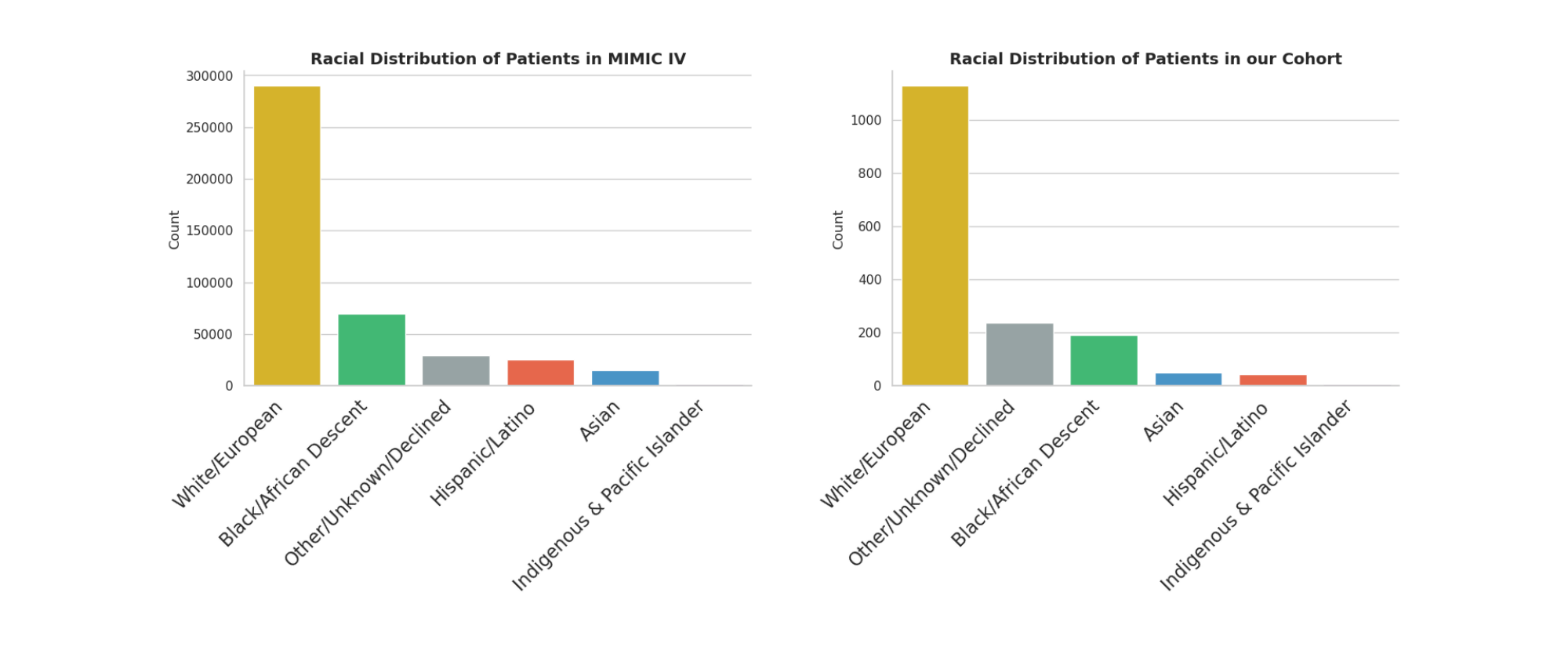}
    \caption{Distribution of patient race in MIMIC IV and our cohort.}
    \label{app:race_mimic_cohort}
\end{figure}

Figure~\ref{app:age_mimic_cohort}~\ref{app:gender_mimic_cohort}~\ref{app:race_mimic_cohort}, present the demographic distributions of patients in MIMIC-IV compared to those in our cohort. The age distribution in MIMIC-IV exhibits a younger mean age compared to our cohort, indicating an older population in our cohort. Gender distribution in our cohort shows a predominance of male patients. The racial distribution of both MIMIC-IV and our cohort shows that they are predominantly composed of White/European patients followed by Black/African Descent patients.

\begin{figure}[h]
    \centering
    \includegraphics[width=0.7\linewidth]{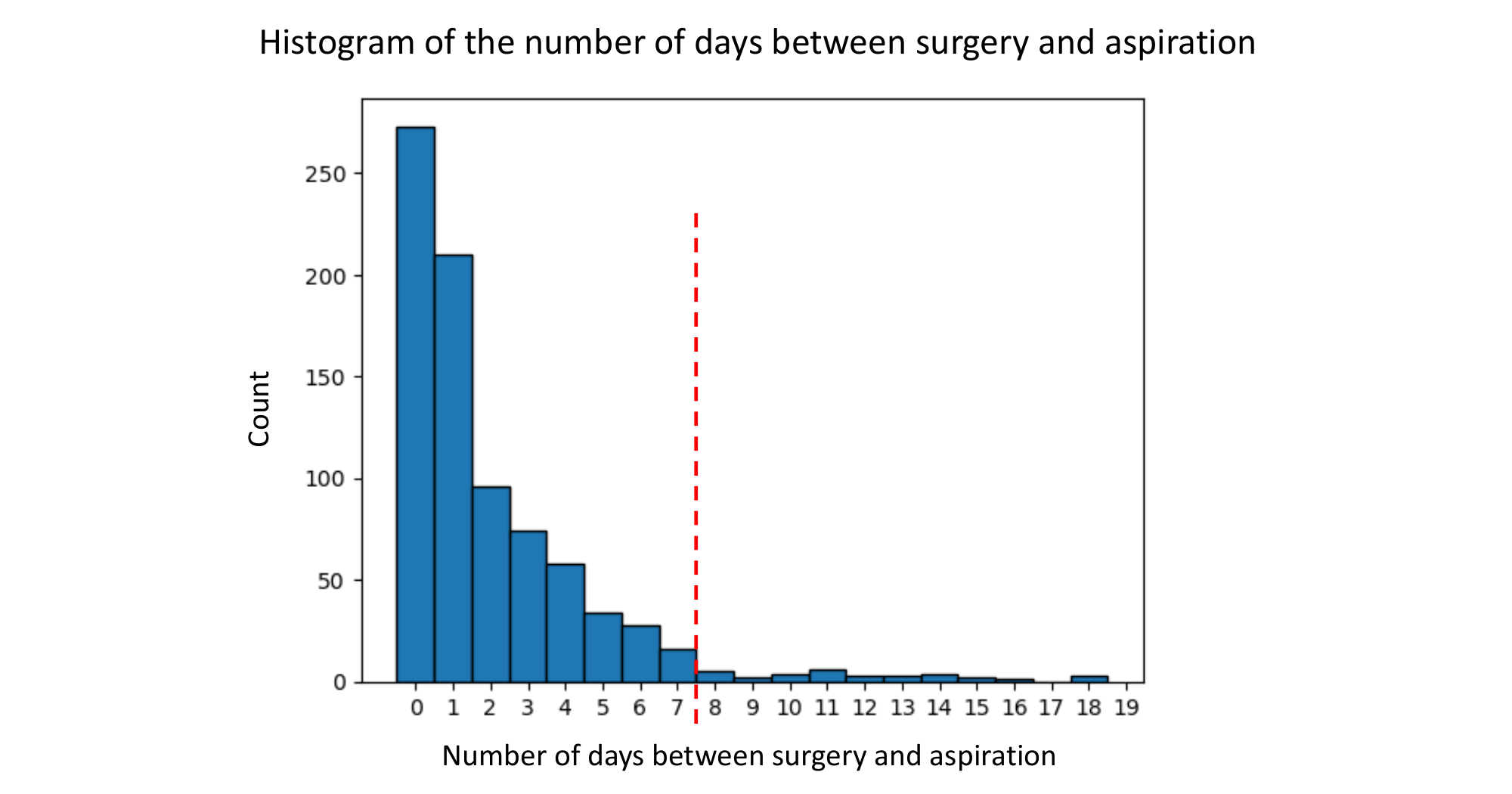}
    \caption{95.6\% of all the aspirations happen within 7 days post operation. The focus of this paper is to predict aspirations within 7 days.}
    \label{fig:postopdays}
\end{figure}

Additionally, we visualize the distribution of days between surgery and aspiration for patients in our cohort, as illustrated in Figure~\ref{fig:postopdays}. Our analysis reveals that the highest number of aspirations occur on the day of surgery, with the incidence steadily declining over time. Notably, more than 95\% of aspirations in our dataset take place within the first seven postoperative days, establishing a rationale for predicting aspiration risk within this timeframe.   

\begin{figure}[h]
    \centering
    \includegraphics[width=0.5\linewidth]{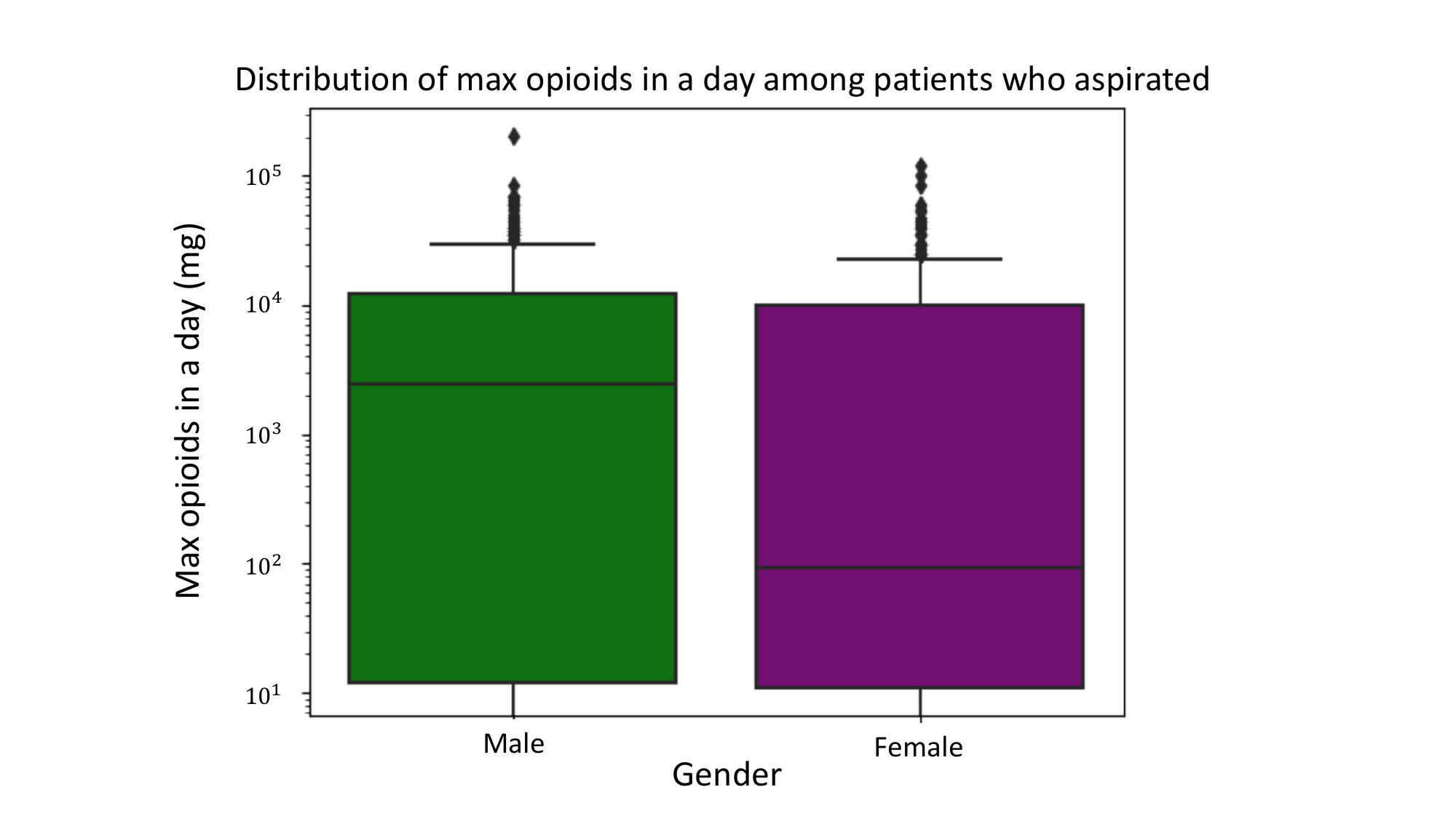}
    \caption{Maximum opioids dosage administered to the patients who aspirated visualized by gender. }
    \label{fig:opioids_gender}
\end{figure}

Finally, Figure~\ref{fig:opioids_gender} presents the maximum MME opioids administered from admission to surgery for patients who experienced postoperative aspiration.

Figure~\ref{fig:surggender} below illustrates the percentage of surgeries performed on different operative sites among patients who aspirated postoperatively. The distribution pattern appears similar between males and females, with no substantial differences observed. In both groups, thoracic procedures were the most frequently. 

\begin{figure}
    \centering
    \includegraphics[width=0.8\linewidth]{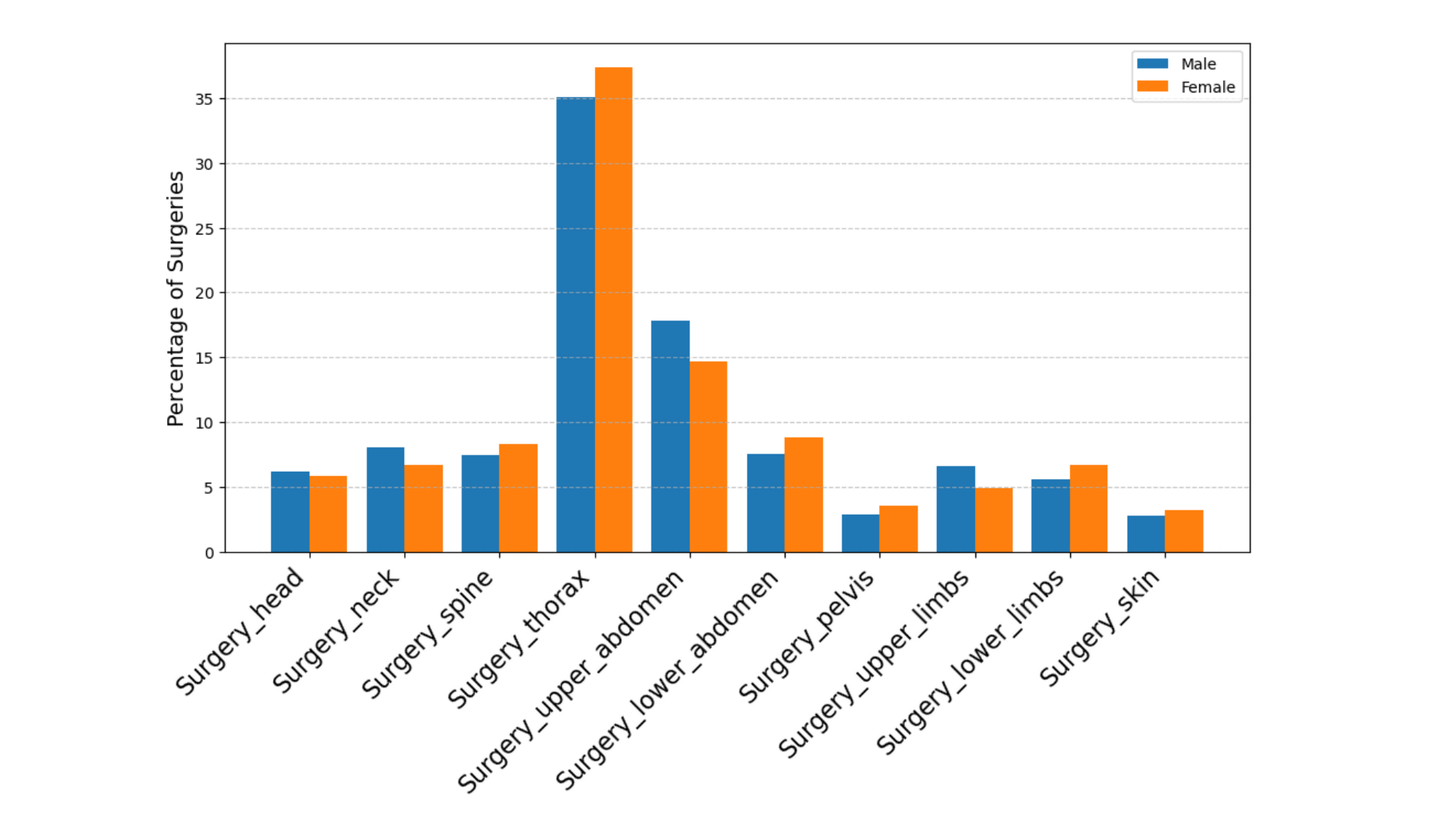}
    \caption{This histogram illustrates the percentage of procedures conducted on each operative site for male and female patients who aspirated postoperatively.}
    \label{fig:surggender}
\end{figure}

\pagebreak

\subsection{Predictive Modeling}

\begin{figure}[h]
    \centering
    \includegraphics[width=0.9\linewidth,trim=0 1.5cm 0 1cm, clip=True]{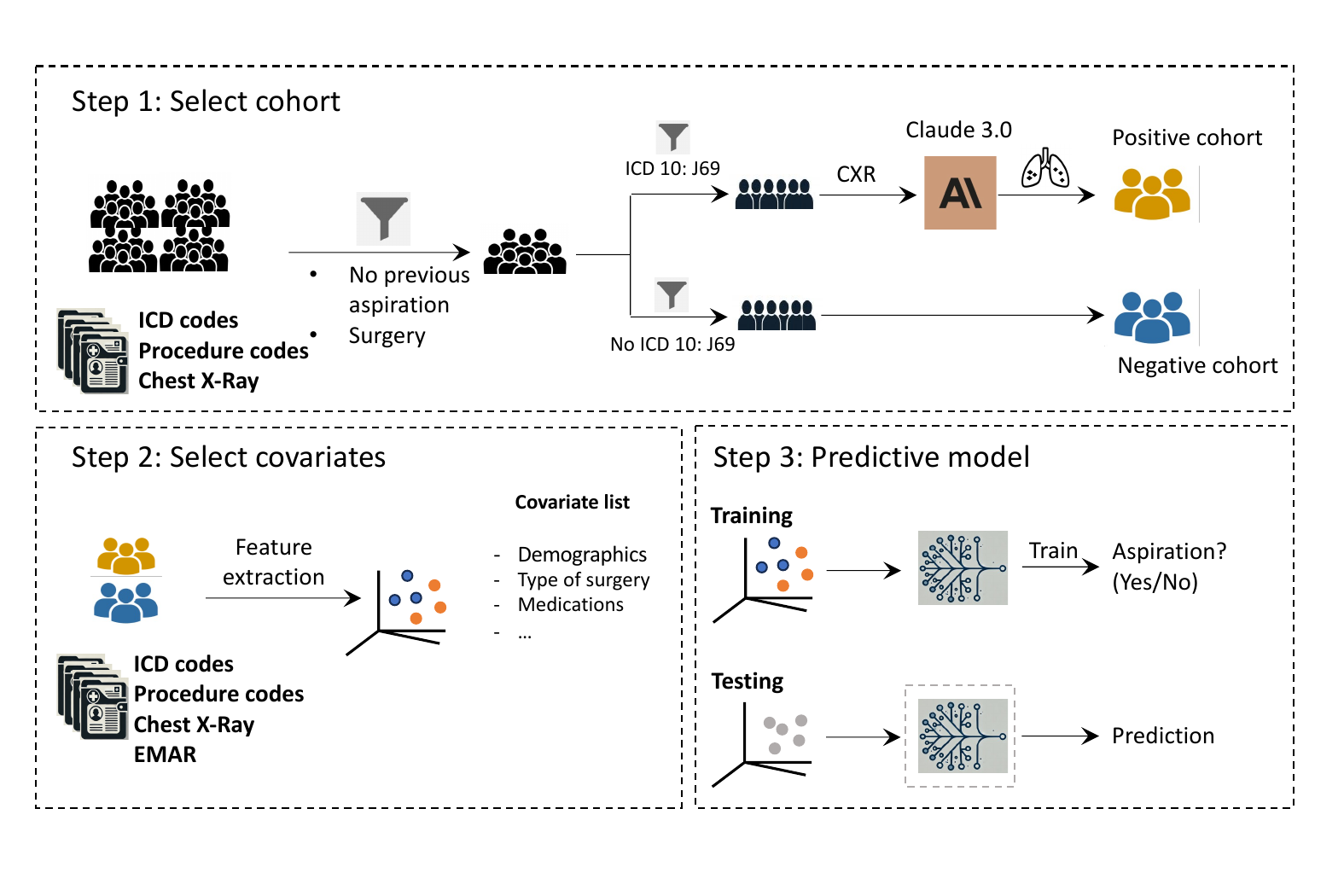}
    \caption{Steps in building a predictive model for post-operative aspiration. 1. We select a cohort of patients who do not have a history of aspiration and underwent surgery. We obtain a positive and negative cohort using the diagnosis of aspiration through ICD code, and the presence of aspiration in the chest X-ray report. 2. From the set of patients identified, we obtain features from their EHR. See Table~\ref{tab:features} for the list of features used in this analysis. 3. Training a model to predict post-operative aspiration given the positive and negative cohort. }
    \label{fig:system_pipeline}
\end{figure}

The methodology used to develop a predictive model for postoperative aspiration risk is illustrated in Figure~\ref{fig:system_pipeline}. The framework consists of three key steps: cohort selection based on clinical criteria, extraction of relevant features from structured medical data, and training a machine learning model to predict aspiration events. This process integrates diagnostic codes, chest X-ray records, medication, and other structured data to construct a comprehensive dataset for model development and evaluation. 

\paragraph{Results.}

\begin{figure}[h]
    \centering
    \includegraphics[width=0.6\linewidth,trim=0 0.8cm 0 1cm, clip=True]{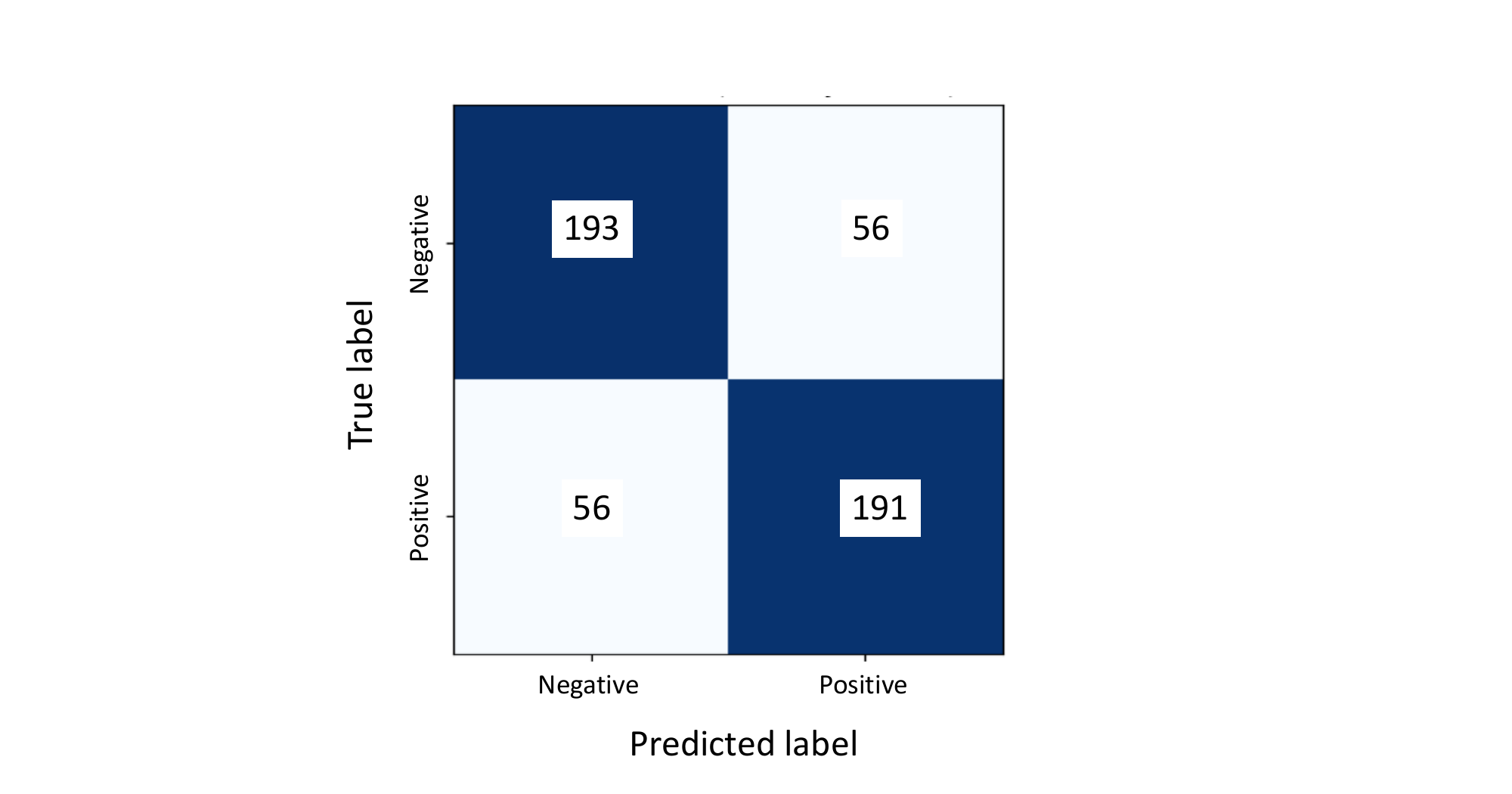}
    \caption{Confusion matrix for predicting post-surgical aspiration with a Random Forest model. Accuracy = 77.42\%. Sensitivity = 77.32\%. Specificity = 77.51\%. AUROC = 0.86}
    \label{fig:confmat}
\end{figure}

Among the models tested, the Random Forest classifier demonstrated superior performance, achieving an overall accuracy of 77.42\% and an area under the receiver operating curve (AUROC) of 0.86. The results of the three different models are shown in Table~\ref{tab:model_accuracies}. The performance of the Random Forest classifier is visualized in the confusion matrix presented in Figure~\ref{fig:confmat}, which highlights the model's ability to correctly classify both aspirating and non-aspirating patients. The model achieves a sensitivity of 77.32\% and specificity of 77.51\%. This result indicates that the model effectively captures the patterns and relationships within the dataset, making it the most reliable choice among the evaluated approaches and demonstrating its potential utility in clinical decision-making. 

\begin{figure}[h]
    \centering
    \includegraphics[width=0.9\linewidth,trim=0 1cm 0 1cm, clip=True]{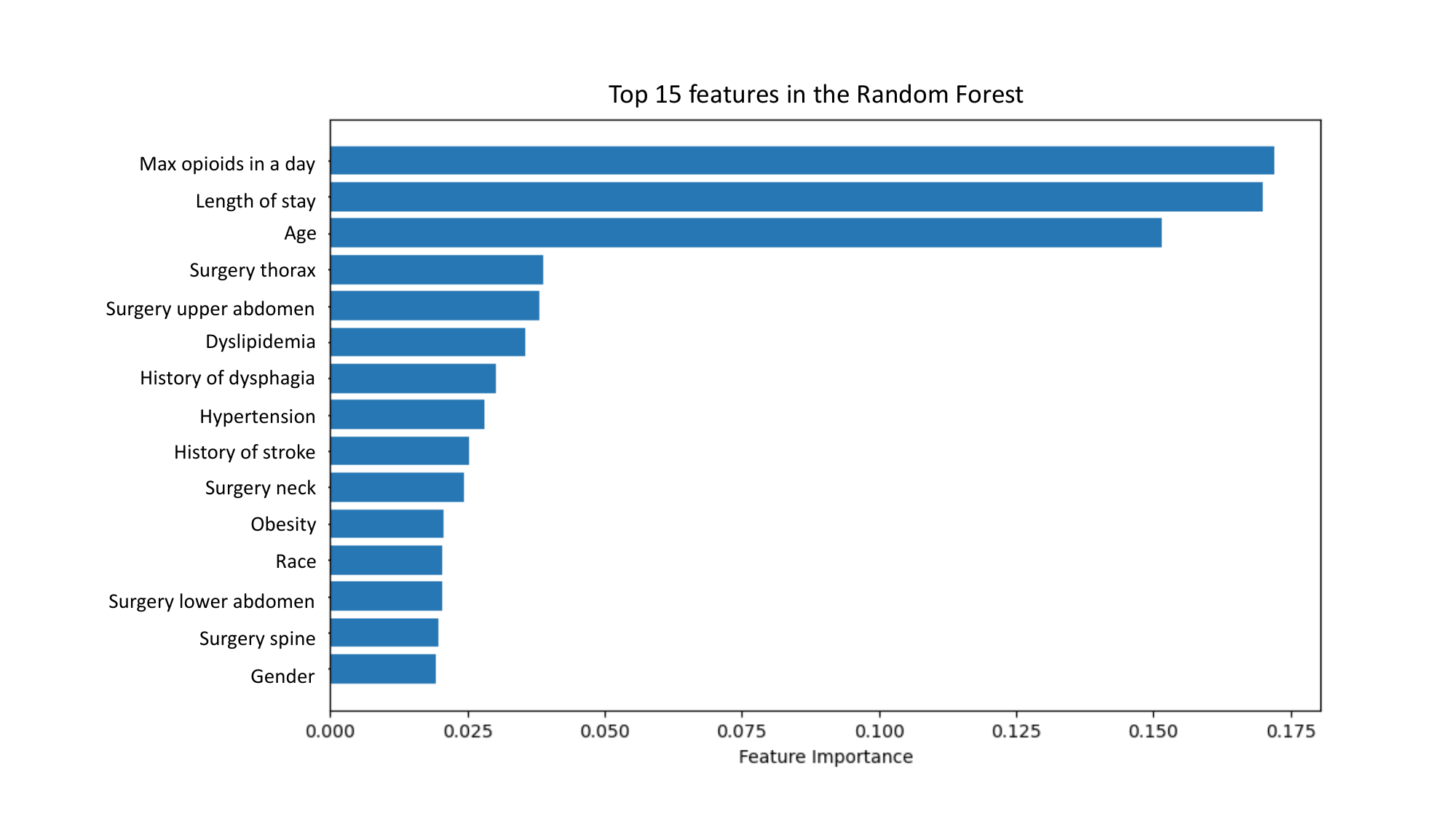}
    \caption{Feature importance scores for the top 15 predictors identified by the Random Forest model, illustrated in decreasing order. The most significant features include maximum opioid dosage in a day, length of hospital stay, patient age, surgeries on the thorax, and surgeries on the upper abdomen.
    }
    \label{fig:feat_imp}
\end{figure}

To better understand the factors driving the model's predictions, feature importance scores were calculated using the Random Forest classifier. The contribution of each feature to the model's ability to reduce uncertainty (measured as the mean decrease in impurity, also known as Gini importance) was quantified across all trees in the ensemble. The top 15 features with the highest importance scores are visualized in Figure~\ref{fig:feat_imp}. Key predictors of aspiration risk, including the maximum daily opioid dosage, the length of hospital stay prior to surgery, age, and surgical location (such as the thorax and upper abdomen), were identified. These findings align with those reported in~\cite{canet2010prediction}, in which interthoracic and upper abdominal surgeries were recognized as risk factors for postoperative pulmonary complications. Insights into potential risk factors are provided by these findings, which could inform clinical decision-making and enable healthcare providers to identify and monitor patients at higher risk of postoperative aspiration more effectively.

\begin{table}[h!]
\centering
\begin{tabular}{c c}
\toprule
\textbf{Model} & \textbf{Accuracy} \\
\toprule
XGBoost & 74.19\% \\
MLP & 76.01\% \\
Random Forest & 77.42\% \\
\bottomrule
\end{tabular}
\caption{\textbf{Aspiration prediction accuracy using different models}. The choice of the hyperparameters used are specified in Section~\ref{sec:hyp}.}
\label{tab:model_accuracies}
\end{table}

\paragraph{Large Language Model Prompt used for Cohort Extraction.}

\label{sec:prompt}

In this work, we utilize Claude 3.0 Sonnet to analyze a patient's chest X-ray (CXR) to determine if there is evidence of aspiration, which is then recorded as the time of aspiration. This is the prompt used for the task:
\begin{tcolorbox}[colback=gray!10, colframe=gray!50, sharp corners, boxrule=0.5mm]
\texttt{You are given a Chest X-Ray of a patient.\\
Your job is to determine if the patient aspirated.\\
Return an answer 0/1 within XML tags <answer></answer>.\\
Return 1 if the patient aspirated and 0 if you think the patient did not aspirate.\\
Do not provide any additional commentary.\\ 
Here is the Chest X-Ray: <report>CXR</report>.}
\end{tcolorbox}

\subsection{Prediction Error Analysis}
\label{sec:erroranalysis}
In predictive modeling, two types of errors are common: false positives and false negatives. For this study, a positive outcome refers to a postoperative aspiration. Among the two, false negatives are significantly more dangerous, as they represent cases where a patient aspirated, but the model failed to predict it accurately. This can lead to a lack of necessary interventions, increasing the risk to the patient's health.



\begin{figure}[h]
    \centering
    \includegraphics[width=0.8\linewidth, trim=0 4.5cm 0 5cm, clip=True]{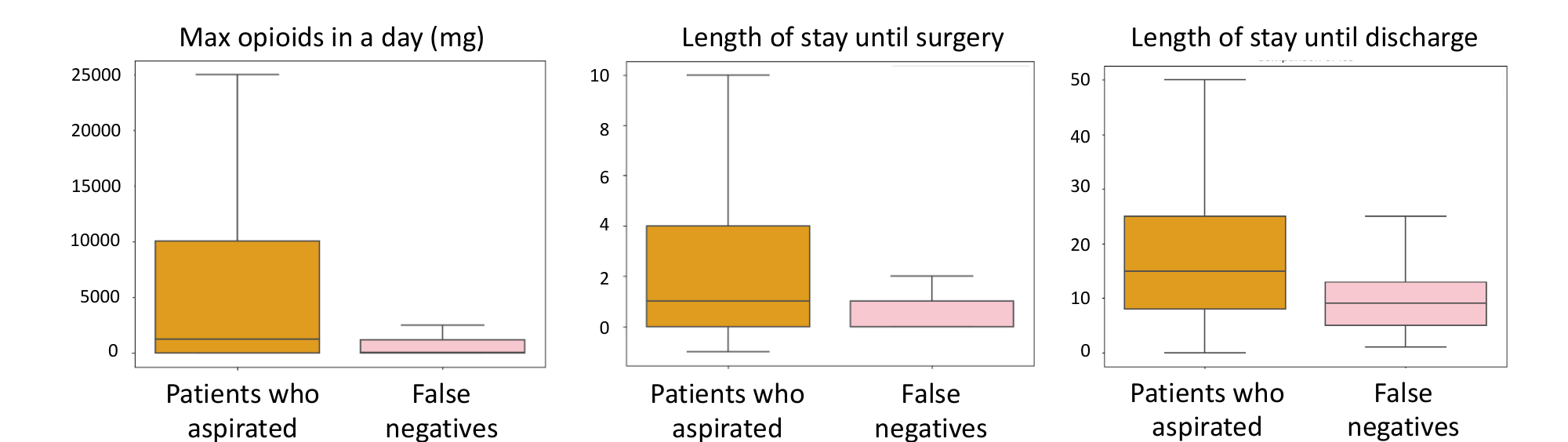}
    \caption{Distribution of maximum daily opioid dosage, length of stay until surgery, and length of stay until discharge (in days) among patients who aspirated and those classified as false negatives—patients who aspirated but were not identified by the predictive model. The analysis shows that false negatives tend to receive lower opioid dosages and have shorter hospital stays compared to the overall distribution of patients who aspirated.}
    \label{fig:diff_falseneg}
\end{figure}

\paragraph{Comparison of False Negatives and Patients who Aspirated.}
Figure~\ref{fig:diff_falseneg} highlights key differences between patients who aspirated and those who aspirated but were misclassified as non-aspirating (false negatives). False-negative patients tend to have lower maximum daily opioid dosages and shorter hospital stays, both until surgery and until discharge. These features, which exhibit high importance in the model's feature ranking, seem to influence the prediction outcomes. A lower opioid dosage and reduced hospitalization time might contribute to reduced model confidence in classifying these patients correctly, potentially because the model associates higher dosages and prolonged stays with aspiration risk.

\begin{figure}[h]
    \centering
    \includegraphics[width=0.6\linewidth,trim=0 2cm 0 4cm, clip=True]{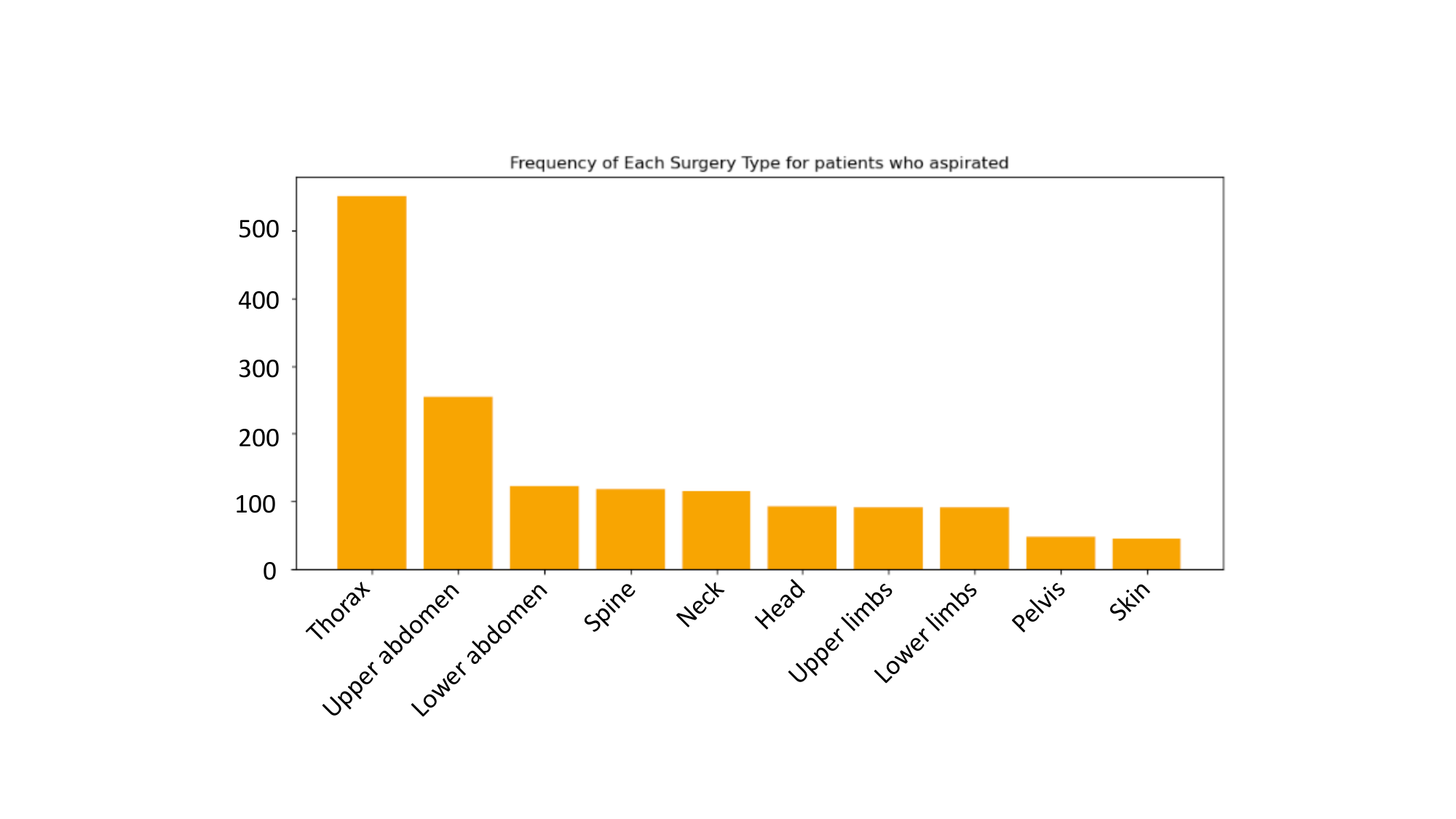}
    \caption{Frequency of each type of surgery among the patients who aspirated.}
    \label{fig:surg_hist_pos}
\end{figure}

\begin{figure}[h]
    \centering
    \includegraphics[width=0.6\linewidth,trim=0 2cm 0 4cm, clip=True]{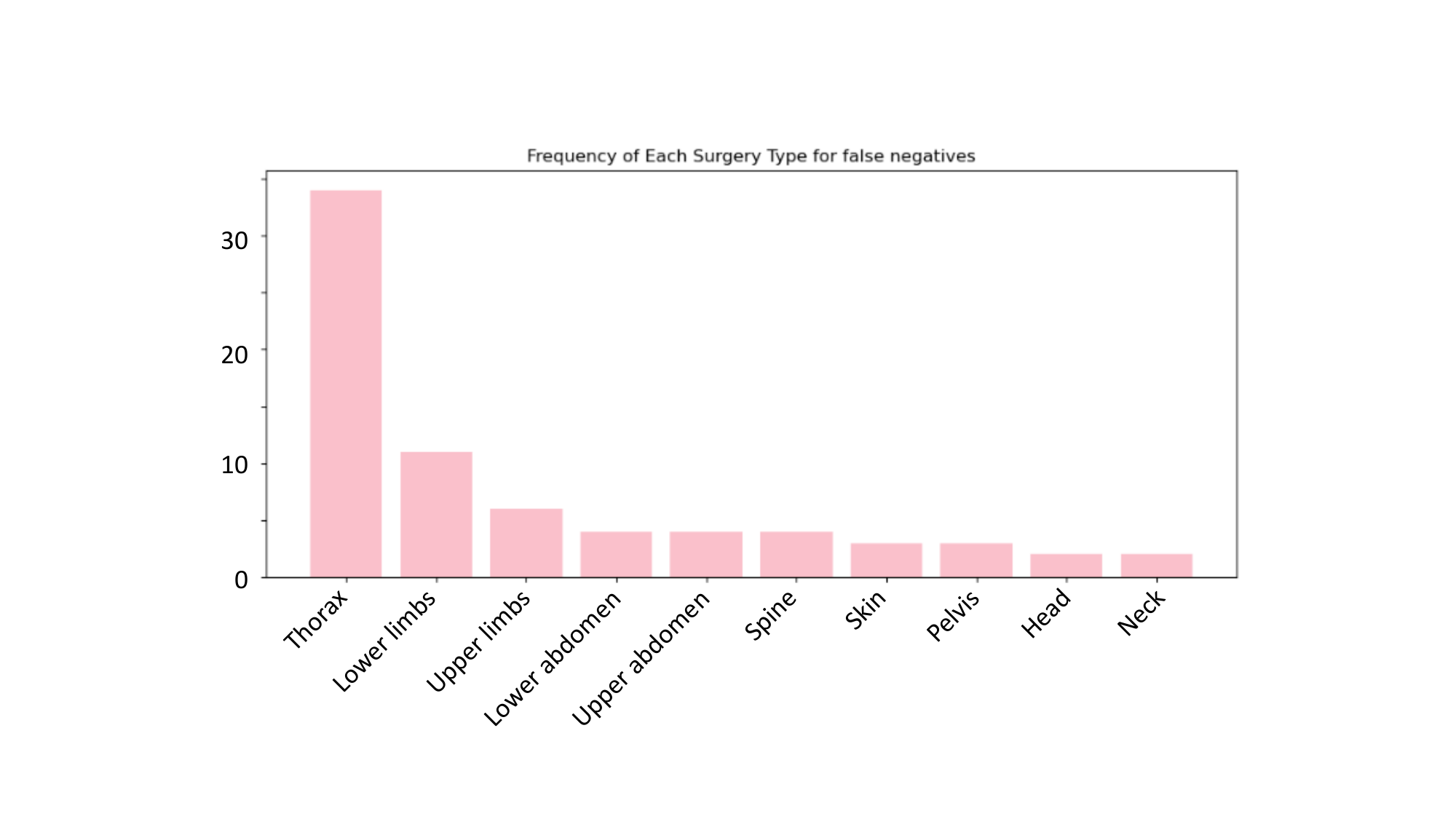}
    \caption{Frequency of different surgeries among false negatives—patients who aspirated but were not detected by the predictive model. Surgeries involving the head and neck, which are more common in the overall distribution of patients who aspirated, appear to be less frequent among the false negatives.}
    \label{fig:surg_hist_falseneg}
\end{figure}

\paragraph{Surgical Profiles of False Negatives.} Figures~\ref{fig:surg_hist_pos} and~\ref{fig:surg_hist_falseneg} compare the distribution of surgery types among patients who aspirated versus those of false negatives, respectively. Thoracic surgeries remain the most common type across both groups, indicating a shared risk factor. However, for false negatives, less frequent surgeries, such as those involving the skin, pelvis, and head, are ranked higher in the distribution. This suggests that the model might struggle with accurately identifying aspiration risks in patients undergoing these less common surgeries, possibly due to a lack of sufficient data representation during training. The rarity of these surgery types in the training data could lead to the model underestimating their association with aspiration risk.


\paragraph{Potential Causes of Errors.} The observed patterns indicate that certain features and underrepresented subgroups in the dataset might contribute to false negatives. Patients who aspirated but had low opioid dosages or underwent uncommon surgeries appear to be particularly challenging for the model to classify. This suggests a need for better representation of these subgroups in the training dataset, as well as a deeper exploration of feature interactions that might improve model sensitivity in detecting such cases.

An interesting observation emerged during the analysis of false negative errors in aspiration prediction. We reviewed the medical report notes that describe the complete hospital admission course of patients classified as false negatives. To streamline the analysis, we utilized a large language model, Claude 3.0 Sonnet. The prompt used for this is provided in Section~\ref{sec:prompt}. Notably, approximately 10 patients had undergone some form of pain control intervention that did not fall under our category of opioids during the feature extraction process from the Electronic Medication Administration Record (EMAR). This exclusion of certain interventions in the feature set may have contributed to the misclassification of these patients, emphasizing the importance of capturing a broader range of pain control measures in future model iterations.



\pagebreak

\paragraph{Large Language Model Prompt used for Error Analysis.}
We use Claude to investigate errors made by the predictive model, particularly focusing on false negatives—patients who aspirated but were missed by the model. To achieve this, we analyze the medical notes associated with these cases, which are often several pages long, and specifically extract information about pain control interventions performed during the patient's admission. 

\begin{tcolorbox}[colback=gray!10, colframe=gray!50, sharp corners, boxrule=0.5mm]
\texttt{You are given a medical report of a patient.\\
Your task is to determine if the patient was given a pain control intervention.\\
Return an answer 0/1 within XML tags <answer></answer>.\\
In addition, return the extract from the note that made you think the patient received pain intervention within <pain></pain>.\\
If the patient did not receive any pain intervention, return <answer>0</answer> <pain>None</pain>.\\
Remember, if your answer is 1, you must provide an extract to justify it.\\
Here is the report: <report>Med report</report>.}
\end{tcolorbox}

\subsection{Hyperparameters and Software Packages Used}
\label{sec:hyp}
\paragraph{Aspiration Prediction Models.}
The Random Forest model was trained using the default hyperparameters provided by the scikit-learn implementation. Specifically, the model used 100 estimators (\texttt{trees}) with the Gini impurity criterion for splitting nodes. The minimum number of samples required to split an internal node was set to 2 (\texttt{min\_samples\_split=2}), and each leaf node was required to contain at least one sample (\texttt{min\_samples\_leaf=1}). Additionally, the maximum number of features considered at each split was the square root of the total number of features (\texttt{max\_features=`sqrt'}).

The Multi-Layer Perceptron (MLP) classifier was configured with a single hidden layer containing 250 neurons (\texttt{hidden\_layer\_sizes=(250,)}). The model used the ReLU activation function (\texttt{activation=`relu'}) and the Adam optimizer (\texttt{solver=`adam'}) for training. The maximum number of iterations for training was set to 500 (\texttt{max\_iter=500}), and a fixed random seed (\texttt{random\_state=42}) was used to ensure reproducibility of results.

The XGBoost model was trained using the \texttt{XGBClassifier} implementation from the \texttt{xgboost} library, with specific hyperparameters chosen for binary classification tasks. 
The objective function was set to logistic regression (\texttt{objective=`binary:logistic'}), and the evaluation metric to minimize during training was log loss (\texttt{eval\_metric=`logloss'}). 
Additionally, a fixed random seed (\texttt{random\_state=42}) was used to ensure reproducibility of the results. All other parameters were left at their default values, providing a robust and general-purpose configuration for binary classification.

\paragraph{ATE Estimation using AIPW.}
In this implementation, the propensity score is estimated using a Decision Tree Classifier. The outcome regression models are Decision Tree Regressors, both from the \texttt{scikit-learn} library with default parameters. Propensity scores are clipped to the range [0.001, 0.999] to ensure numerical stability. The bootstrap procedure uses 1,000 resampling iterations to calculate confidence intervals for the ATE estimate.

\end{document}